\documentclass{article}
\usepackage[nonatbib]{main}

\usepackage{hyperref}

\usepackage{amsthm}
\usepackage{amsmath,amssymb}
\usepackage{enumitem}

\usepackage[sort&compress,numbers]{natbib}
\usepackage[normalem]{ulem}

\usepackage{times}
\usepackage{graphicx} 

\graphicspath{{../fig/}}

\usepackage{tikz}
\usepackage{tkz-tab}
\usepackage{caption} 
\usepackage{subcaption} 
\usetikzlibrary{shapes.geometric, arrows}
\tikzstyle{arrow} = [very thick,->,>=stealth]

\usepackage{cleveref}
\usepackage{setspace}
\usepackage{wrapfig}
\usepackage[ruled]{algorithm}
\usepackage{algpseudocode}

\usepackage[disable]{todonotes}

\usepackage{physics}
\usepackage{bm}
\usepackage{mathtools}
\usepackage{siunitx}
\usepackage{booktabs}
\usepackage{multirow}
\usepackage{colortbl}
\usepackage{tabularx}
\usepackage{adjustbox}
\usepackage{algorithm}
\usepackage{algpseudocode}

\DeclarePairedDelimiterX{\infdivx}[2]{(}{)}{%
  #1\;\delimsize\|\;#2%
}

\DeclareMathOperator*{\argmin}{arg\,min}

\title{A Discussion on Solving Partial Differential Equations using Neural Networks}

\author{
	Tim Dockhorn \\
	Department of Applied Mathematics\\
	University of Waterloo\\
	Waterloo, ON, N2L 3G1 \\
	\texttt{tim.dockhorn@uwaterloo.ca} \\
}

\begin{document}
\maketitle

\begin{abstract} 
Can neural networks learn to solve partial differential equations (PDEs)? We investigate this question for two (systems of) PDEs, namely, the Poisson equation and the steady Navier--Stokes equations. The contributions of this paper are five-fold. (1) Numerical experiments show that small neural networks ($< 500$ learnable parameters) are able to accurately learn complex solutions for systems of partial differential equations. (2) It investigates the influence of random weight initialization on the quality of the neural network approximate solution and demonstrates how one can take advantage of this non-determinism using ensemble learning. (3) It investigates the suitability of the loss function used in this work. (4) It studies the benefits and drawbacks of solving (systems of) PDEs with neural networks compared to classical numerical methods. (5) It proposes an exhaustive list of possible directions of future work.
\end{abstract} 

\section{Introduction}
Partial differential equations can be used to model a vast variety of phenomena in areas of natural sciences as well as engineering and finance. Even though most PDEs do not have an analytical solution, their solution can be approximated using classical numerical methods (which are based on a discretization of the domain). These methods are particularly efficient for low-dimensional problems on regular geometries; however, finding an appropriate discretization for a complex geometry can be as difficult as solving the partial differential equation itself. This problem is particularly severe if the space dimension is large as there is no straightforward way to discretize irregular domains in space dimensions larger than three. 

Advantages in deep learning have lead to remarkable performance both in computer vision and natural language processing using convolutional neural networks \cite{szegedy2015going, he2016deep} and recurrent neural networks \cite{graves2013speech, mikolov2011extensions}, respectively. Recently, work has been proposed for learning partial differential equations using neural networks \cite{sirignano2018dgm, raissi2017physics, berg2018unified}. The contributions of this paper are five-fold:
\begin{enumerate}
    \item We show that small neural networks are able to accurately learn the solution for an instance of the Poisson equation as well as an instance of the Navier--Stokes equations.
    \item We demonstrate that there is a non-negligible influence of the random weight initialization on the approximate solution of the neural network and propose to use ensemble methods to exploit this non-determinism.
    \item The used loss functions, which can be seen as a penalty method from an optimization point of view, have no theoretical justification. We attempt to empirically advocate the loss functions by measuring the correlation of two quantities. Different intensities of correlation can be seen for the Poisson problem and the Navier--Stokes problem. 
    \item We study the benefits and drawbacks of solving (systems of) PDEs with neural networks compared to classical numerical methods.
    \item We propose a vast variety of topics that could stimulate future work in the field of solving (systems of) PDEs using neural networks.
\end{enumerate}
All code and datasets to reproduce the results of this work are available at~\url{https://github.com/timudk/SPDENN}.

\section{Algorithm}
In this section, we review the algorithm used in \cite{sirignano2018dgm, raissi2017physics} to learn the solution of a steady partial differential equation. Let $\Omega \subset \mathbb{R}^d$ be a domain and let $\partial \Omega \subset \mathbb{R}^{d-1}$ denote the boundary of $\Omega$. Consider the following PDE subject to Dirichlet boundary conditions
\begin{align} \label{eq:general_pde}
\begin{split}
    \mathcal{N} u(\bm{x}) &= f(\bm{x}) \quad \text{in } \Omega, \\
    u(\bm{x}) &= g(\bm{x}) \quad \text{on } \partial \Omega,
\end{split}
\end{align}
where $d$ is the number of space dimensions. $\mathcal{N}$ is a differential operator and $f$ as well as $g$ are known right-hand-sides. The algorithm approximates $u(\bm{x})$ by a neural network $\hat{u}(\bm{x}, \bm{\theta})$, where $\bm{\theta}$ is the stacked vector of the neural network parameters. Both, \cite{sirignano2018dgm} and \cite{raissi2017physics}, propose a loss function of the form 
\begin{align} \label{eq:genereic_loss}
    L_{MC}(\bm{\theta}) = \frac{1}{N_{\text{int}}} \sum_{i=1}^{N_{\text{int}}} \left(\mathcal{N}\hat{u}(\bm{x}_i) - f(\bm{x}_i)\right)^2 + \frac{1}{N_{\text{bou}}} \sum_{j=1}^{N_{\text{bou}}} \left(\hat{u}(\bm{s}_j)-g(\bm{s}_j)\right)^2,
\end{align}
where the elements of $\{\bm{x}_i\}_{i=1}^{N_{\text{int}}}$ and $\{\bm{s}_j\}_{i=1}^{N_{\text{bou}}}$ are uniformly sampled from $\Omega$ and $\partial \Omega$, respectively. 

From an optimization point of view, we converted the constrained problem~\eqref{eq:general_pde} into an unconstrained problem using a penalty method for a finite number of constraints $\hat{u}(\bm{s}_j) = g(\bm{s}_j) \; \forall j = 1, \cdots, N_{\text{bou}}$ (instead of the original constraint $\hat{u}(\bm{x}) = g(\bm{x})$ on $\partial \Omega$) and penalty coefficients equal to $1/N_{\text{bou}}$; we refer the reader to~\citet{nocedal2006numerical} for more information on penalty methods. Note that equation~\eqref{eq:genereic_loss} is a Monte Carlo approximation of 
\begin{align}
    L(\bm{\theta})  = \mathop{\mathbb{E}}_{\bm{x} \sim \mathcal{U}(\Omega)} \left[ \left( \mathcal{N} \hat{u}(\bm{x}) - f(\bm{x}) \right)^2 \right] + \mathop{\mathbb{E}}_{\bm{s} \sim \mathcal{U}(\partial \Omega)} \left[ \left( \hat{u}(\bm{s}) - g(\bm{s}) \right)^2 \right],
\end{align}
where $\mathcal{U}(\Omega)$ and $\mathcal{U}(\partial \Omega)$ denote the uniform distribution on the domain and the boundary, respectively. This algorithm can be readily extended to systems of partial differential equations by summing up the loss functions of each equation. Note that no knowledge of the true solution $u(\bm{x})$ is needed.
\section{Experiments} \label{se:experiments}
In this work, we use the Broyden--Fletcher--Goldfarb--Shanno (BFGS) algorithm to optimize the network parameters $\bm{\theta}$. We further use the sigmoid activation function in combination with Xavier initialization \cite{glorot2010understanding}. 

We train neural networks for two problems, the Poisson problem and the steady Navier--Stokes problem; for both problems we know the true solution and can therefore measure the quality of the approximate solution. In order to avoid any errors introduced by numerical integration, we measure the quality of an approximate solution in a heuristic inspired by the finite difference method~\cite{larsson2008partial} instead of using the $L^2$-norm for which integrals have to be evaluated. Let $v$ be a known function and $\hat{v}$ be an approximation to $v$, we then measure the quality of $\hat{v}$ as 
\begin{align}
    FD(v - \hat{v}) = \frac{1}{N_{\text{meas}}} \sum_{i=1}^{N_{\text{meas}}} \left(v(\bm{x}_i)-\hat{v}(\bm{x}_i) \right)^2 \approx \mathop{\mathbb{E}}_{\bm{x} \sim \mathcal{U}(\Omega)} \left[ \left( v(\bm{x}) - \hat{v}(\bm{x}) \right)^2 \right],
\end{align}
where $\{\bm{x}\}_{i=1}^{N_{\text{meas}}}$ is a set of uniformly spaced grid points over $\Omega \subset \mathbb{R}^d$. Since $\Omega$ is a rectangle for both considered problems, the grid points are chosen in a way such that two neighboring points have the same value in $d-1$ coordinates and only differ by $0.01$ in one coordinate.

\subsection{Poisson equation}
\begin{table}
    \caption{Numerical results for the manufactured Poisson problem with loss $\sqrt{L_{MC}^{\text{p}}}$.}
    \label{tb:poisson_1}
    \centering
    \begin{adjustbox}{tabular=lllllll,center}
    \toprule
    Dataset & \multicolumn{2}{c}{1} & \multicolumn{4}{c}{2} \\
    \cmidrule(r){1-1} \cmidrule(r){2-3} \cmidrule(r){4-7} 
    \# Hidden layers & 1 & 2 & 3 & 4 & 3 & 4 \\
    \midrule
    $\sqrt{FD(u-\hat{u})}$ & 
    \num[round-precision=2,round-mode=figures,scientific-notation=true]{0.00000046968838446439} & 
    \num[round-precision=2,round-mode=figures,scientific-notation=true]{0.00000177130415045161} &
    \num[round-precision=2,round-mode=figures,scientific-notation=true]{0.000000842502273212102} &
    \num[round-precision=2,round-mode=figures,scientific-notation=true]{0.0000013142280798682} &
    \num[round-precision=2,round-mode=figures,scientific-notation=true]{0.0000102947287092007} &
    \num[round-precision=2,round-mode=figures,scientific-notation=true]{0.00000709839066140018} \\
    $\sqrt{FD(\nabla^2 \hat{u} + f)}$ &
    \num[round-precision=2,round-mode=figures,scientific-notation=true]{0.0000555331012962603} &
    \num[round-precision=2,round-mode=figures,scientific-notation=true]{0.000024511333338069} &
    \num[round-precision=2,round-mode=figures,scientific-notation=true]{0.000063119469898984} &
    \num[round-precision=2,round-mode=figures,scientific-notation=true]{0.0000122215321565039} &
    \num[round-precision=2,round-mode=figures,scientific-notation=true]{0.0000754143002917667} &
    \num[round-precision=2,round-mode=figures,scientific-notation=true]{0.0000484014501604002} \\
    iterations &
    $<13000$ & $<12000$ & $20000$ & $<14000$ & $<8000$ & $<8000$ \\
    \midrule
    Dataset & \multicolumn{4}{c}{3} & \multicolumn{2}{c}{4} \\
    \cmidrule(r){1-1} \cmidrule(r){2-5} \cmidrule(r){6-7} 
    \# Layers & 1 & 2 & 3 & 4 & 3 & 4 \\
    \midrule
    $\sqrt{FD(u-\hat{u})}$ & 
    \num[round-precision=2,round-mode=figures,scientific-notation=true]{0.000000392900991432573} \cellcolor[gray]{.8}& 
    \num[round-precision=2,round-mode=figures,scientific-notation=true]{0.00000173006892599481} &
    \num[round-precision=2,round-mode=figures,scientific-notation=true]{0.00000138276964511926} &
    \num[round-precision=2,round-mode=figures,scientific-notation=true]{0.000000718840952053486} &
    \num[round-precision=2,round-mode=figures,scientific-notation=true]{0.00000129535461791372} &
    \num[round-precision=2,round-mode=figures,scientific-notation=true]{0.00000105370314221945} \\
    $\sqrt{FD(\nabla^2 \hat{u} + f)}$ &
    \num[round-precision=2,round-mode=figures,scientific-notation=true]{0.0000624076985106196} &
    \num[round-precision=2,round-mode=figures,scientific-notation=true]{0.0000142855933184652} &
    \num[round-precision=2,round-mode=figures,scientific-notation=true]{0.0000155849742035369} &
    \num[round-precision=2,round-mode=figures,scientific-notation=true]{0.00000744141762690598} &
    \num[round-precision=2,round-mode=figures,scientific-notation=true]{0.0000121825362913851} &
    \num[round-precision=2,round-mode=figures,scientific-notation=true]{0.0000140683514232559} \\
    iterations &
    $20000$ & $<12000$ & $<9000$ & $<6000$ & $<10000$ & $<6000$ \\
    \bottomrule
    \end{adjustbox}
\end{table}

The Poisson equation in combination with Dirichlet boundary conditions
\begin{align}
\begin{split}
    -\nabla^2 u (\bm{x}) &= f(\bm{x}) \quad \text{in } \Omega, \\
    u (\bm{x}) &= g(\bm{x}) \quad \text{on } \partial \Omega,
\end{split}
\end{align}
is one of the most widely studied boundary value problems and serves well as a first numerical experiment. In this work, we studied the Poisson problem for $d=2$ using neural networks from one up to four hidden layers, with 16 units each. The networks are trained on four datasets with 1000, 2000, 4000, and 8000 interior as well as boundary data points, respectively. For the Poisson problem the loss function can be written as 
\begin{align}
    L_{MC}^{\text{p}}(\hat{u}; \bm{\theta}) = \frac{1}{N_{\text{int}}} \sum_{i=1}^{N_{\text{int}}} \left( \nabla^2 \hat{u}(\bm{x}_i) + f(\bm{x}_i)\right)^2 + \frac{1}{N_{\text{bou}}} \sum_{j=1}^{N_{\text{bou}}} \left(\hat{u}(\bm{s}_j)-g(\bm{s}_j)\right)^2.
\end{align}
We consider the manufactured problem $u=\sin(\pi x_1) \sin(\pi x_2)$, resulting in $f = 2 \pi^2 \sin(\pi x_1) \sin(\pi x_2)$, in the domain $\Omega = \left[ 0, 1\right]^2$. We found that our results could be significantly improved if we use $\sqrt{L_{MC}^{\text{p}}}$ instead of $L_{MC}^{\text{p}}$ as our loss function; this is most likely due to $L_{MC}^{\text{p}} \ll 1$ being too small to effectively update the weights of the neural network. 

Numerical results for the different networks can be found in Table~\ref{tb:poisson_1}. We set the number of maximum BFGS iterations to 20000, however, most simulations terminated early due to a violation of the Wolfe conditions~\cite{wolfe1969convergence}. \citet{berg2018unified} proposed to run a few hundred stochastic gradient descent steps with a stepsize of $10^{-9}$ when the Wolfe conditions are violated, in order to get out of the troublesome region, and then run BFGS again. Due to the high number of networks being trained for this work, we did not consider this approach and stopped the optimization process once the Wolfe conditions were violated.

\begin{figure}
    \begin{subfigure}{.5\textwidth}
      \centering
      \includegraphics[width=\linewidth]{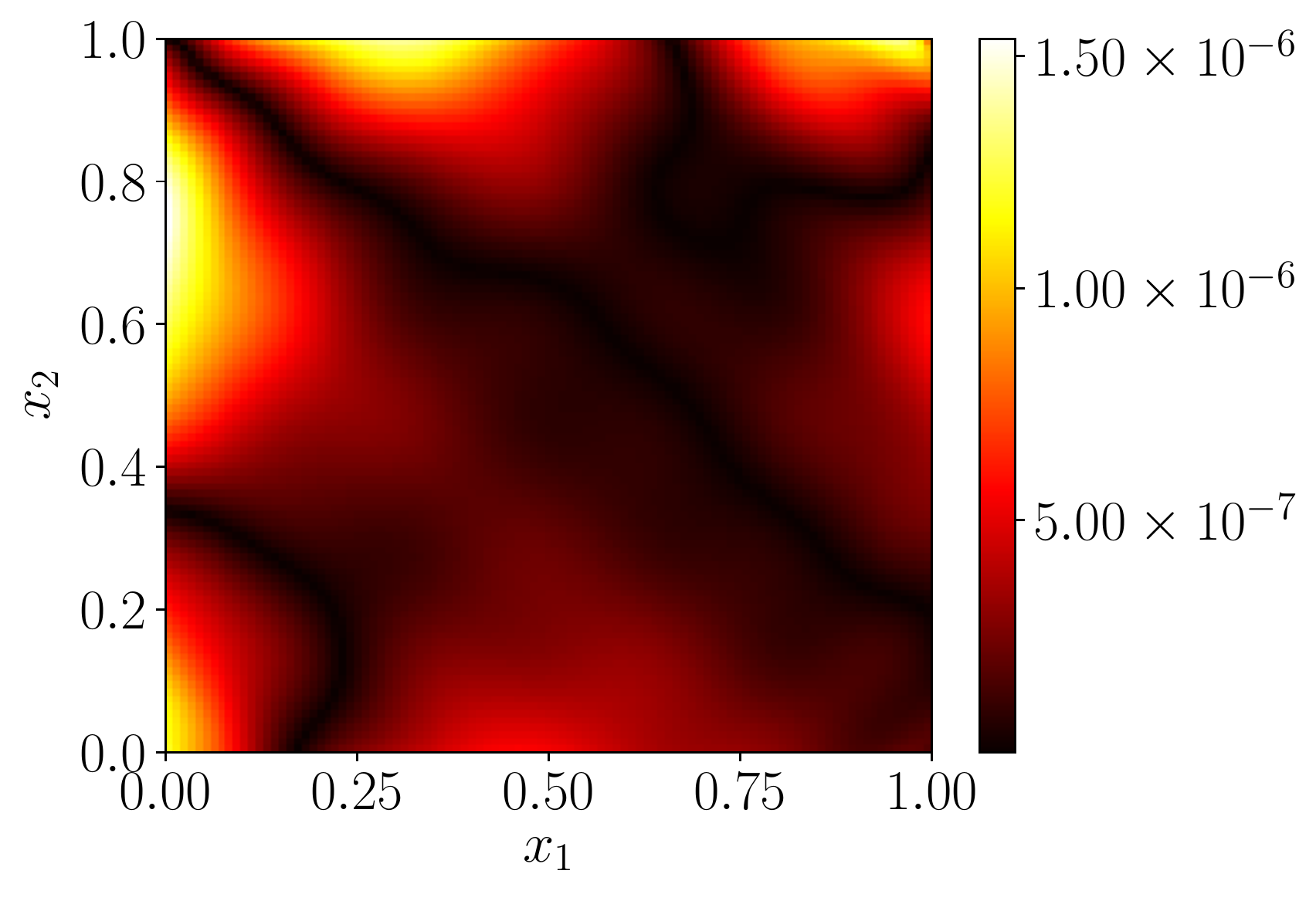}
      \caption{$\abs{u(\bm{x})-\hat{u}(\bm{x})}$}
    \end{subfigure}%
    \begin{subfigure}{.5\textwidth}
      \centering
      \includegraphics[width=\linewidth]{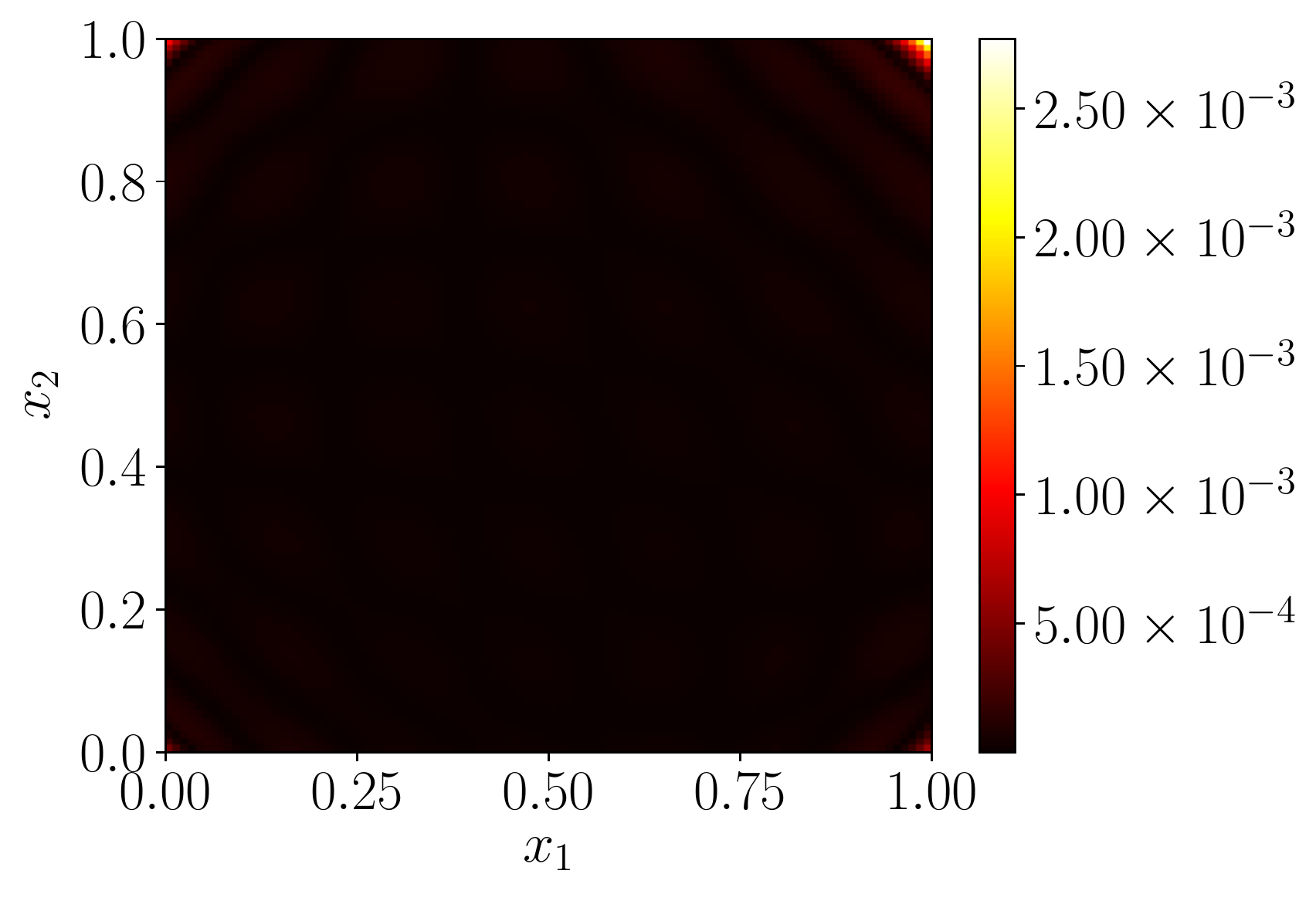}
      \caption{$\abs{\nabla^2 \hat{u}(\bm{x}) + f(\bm{x})}$}
    \end{subfigure}
    \caption{Numerical solution of the manufactured Poisson problem using one hidden layer, dataset 1 and loss $\sqrt{L_{MC}^{\text{p}}}$.}
    \label{fig:poisson}
\end{figure}

Surprisingly, the best result, i.e., the lowest value of $\sqrt{FD(u-\hat{u})}$, is achieved for the neural network with just one hidden layer (using dataset 3); a plot of $\abs{u(\bm{x})-\hat{u}(\bm{x})}$ and $\abs{\nabla^2 \hat{u}(\bm{x}) + f(\bm{x})}$ for this particular network can be found in Figure~\ref{fig:poisson}. Note that the value of $\abs{\nabla^2 \hat{u}(\bm{x}) + f(\bm{x})}$ is particularly high in the top right-hand corner; we attempt to counteract this problem by training some networks with the modified loss function
\begin{align} \label{eq:mod_loss}
    \sqrt{\bar{L}_{MC}^{\text{p}}} = \sqrt{L_{MC}^{\text{p}} + \frac{1}{N_{\text{corner}}} \sum_{i=1}^{N_{\text{corner}}} \left( \nabla^2 \hat{u}(\bm{x}_i) + f(\bm{x}_i)\right)^2 + \frac{1}{N_{\text{corner}}} \sum_{j=1}^{N_{\text{corner}}} \left(\hat{u}(\bm{s}_j)-g(\bm{s}_j)\right)^2},
\end{align}
where $\{\bm{x}_i\}_{i=1}^{N_{\text{corner}}} = \{ (0.0, 0.0), (1.0, 0.0), (0.0, 1.0), (1.0, 1.0) \}$ are the corner points of $\Omega$.

In order to study the influence of the random weight initialization, we train ten neural networks (all having one hidden layer and random seeds from 42 up to 51) each on six combinations of datasets 1-3 and loss functions $\sqrt{L_{MC}^{\text{p}}}$ and $\sqrt{\bar{L}_{MC}^{\text{p}}}$. We then compute arithmetic means $\mu$ and standard deviations $\sigma$ for the quantities $\sqrt{FD(u-\hat{u})}$ and $\sqrt{FD(\nabla \hat{u} + f)}$ over the ten networks; results can be found in Table~\ref{tb:poisson_2}.

The best result for $\mu_{\sqrt{FD(u-\hat{u})}}$, when using $\sqrt{L_{MC}^{\text{p}}}$, is again achieved when dataset 3 is used. Even though using $\sqrt{\bar{L}_{MC}^{\text{p}}}$ leads to extremely small values of $\abs{u(\bm{x})-\hat{u}(\bm{x})}$ and $\abs{\nabla^2 \hat{u}(\bm{x}) + f(\bm{x})}$ in the corners of $\Omega$, the quantity $\mu_{\sqrt{FD(u-\hat{u})}}$ is larger for all three datasets compared to when using the standard loss function $\sqrt{L_{MC}^{\text{p}}}$. 

In general, the standard deviation of both quantities are of the order of the arithmetic means implying that the initialization has a big influence on the result; a smart way of initializing the neural network is therefore an important question for future work.

We further want to investigate whether or not $\sqrt{FD(u-\hat{u})}$ is a good indicator for $\sqrt{FD(\nabla^2 \hat{u} + f)}$, i.e., whether or not a decreasing/increasing value of the former corresponds to a decreasing/increasing value of the latter. To do so we compute the Kendall $\tau$ coefficient \cite{kendall1938new}
\begin{align}
\tau = \frac{(\text{number of concordant pairs) - (\text{number of discordant pairs})}}{n(n-1)/2},
\end{align}
for each combination of dataset and loss function over the ten networks. Examples for a concordant and a discordant pair (of tuples) are $\{(3,4), (1,0)\}$ and $\{(3,4), (4,2)\}$, respectively.
Note that if the agreement between two rankings is perfect, the coefficient $\tau$ has value 1. The Kendall $\tau$ coefficient is $\leq 0.6$ for all combinations of loss functions and datasets (see Table~\ref{tb:poisson_2}, indicating that there is no strong relation between the two quantities. As both loss functions, $\sqrt{L_{MC}^{\text{p}}}$ and $\sqrt{\bar{L}_{MC}^{\text{p}}}$, are strongly related with $\sqrt{FD(\nabla^2 \hat{u} + f)}$, it is unclear how well-suited they are in minimizing the difference between the true solution $u$ and the neural network approximate $\hat{u}$.
\begin{table}[]
    \centering
    \caption{Statistics of numerical results for the manufactured Poisson problem using one hidden layer.}
    \label{tb:poisson_2}
    \begin{adjustbox}{tabular=lllllll,center}
        \toprule
        Dataset & \multicolumn{2}{c}{1} & \multicolumn{2}{c}{2} & \multicolumn{2}{c}{3} \\
        \cmidrule(r){1-1} \cmidrule(r){2-3} \cmidrule(r){4-5} \cmidrule(r){6-7} 
        Loss function & $\sqrt{L_{MC}^{\text{p}}}$  & $\sqrt{\bar{L}_{MC}^{\text{p}}}$ & $\sqrt{L_{MC}^{\text{p}}}$  & $\sqrt{\bar{L}_{MC}^{\text{p}}}$ & $\sqrt{L_{MC}^{\text{p}}}$  & $\sqrt{\bar{L}_{MC}^{\text{p}}}$ \\
        \cmidrule(r){1-1} \cmidrule(r){2-2} \cmidrule(r){3-3} \cmidrule(r){4-4} \cmidrule(r){5-5} \cmidrule(r){6-6} \cmidrule(r){7-7} 
        $\mu_{\sqrt{FD(u-\hat{u})}}$ &
        \num[round-precision=2,round-mode=figures,scientific-notation=true]{0.00000715550947105535} &
        \num[round-precision=2,round-mode=figures,scientific-notation=true]{0.0000104963664185061} &
        \num[round-precision=2,round-mode=figures,scientific-notation=true]{0.0000133316902197093} &
        \num[round-precision=2,round-mode=figures,scientific-notation=true]{0.0000120473243618109} &
        \num[round-precision=2,round-mode=figures,scientific-notation=true]{0.0000055345295833325} \cellcolor[gray]{.8}&
        \num[round-precision=2,round-mode=figures,scientific-notation=true]{0.0000138758522303236} \\
        $
        \sigma_{\sqrt{FD(u-\hat{u})}}$ &
        \num[round-precision=2,round-mode=figures,scientific-notation=true]{0.00000517356969612896} &
        \num[round-precision=2,round-mode=figures,scientific-notation=true]{0.00000630832098152191} &
        \num[round-precision=2,round-mode=figures,scientific-notation=true]{0.0000145393433678106} &
        \num[round-precision=2,round-mode=figures,scientific-notation=true]{0.0000121173598304748} &
        \num[round-precision=2,round-mode=figures,scientific-notation=true]{0.00000508132817564874} &
        \num[round-precision=2,round-mode=figures,scientific-notation=true]{0.0000129089410886698} \\
        $\mu_{\sqrt{FD(\nabla^2 \hat{u} - f)}}$ &
        \num[round-precision=2,round-mode=figures,scientific-notation=true]{0.000988813469524702} &
        \num[round-precision=2,round-mode=figures,scientific-notation=true]{0.000864556565273151} &
        \num[round-precision=2,round-mode=figures,scientific-notation=true]{0.000599062147692847} &
        \num[round-precision=2,round-mode=figures,scientific-notation=true]{0.000853524005336476} &
        \num[round-precision=2,round-mode=figures,scientific-notation=true]{0.000727009438579735} &
        \num[round-precision=2,round-mode=figures,scientific-notation=true]{0.000655774962274615} \\
        $\sigma_{\sqrt{FD(\nabla^2 \hat{u} - f)}}$ &
        \num[round-precision=2,round-mode=figures,scientific-notation=true]{0.000286191280349517} &
        \num[round-precision=2,round-mode=figures,scientific-notation=true]{0.000460365698041587} &
        \num[round-precision=2,round-mode=figures,scientific-notation=true]{0.000248710435219303} &
        \num[round-precision=2,round-mode=figures,scientific-notation=true]{0.000425840741206983} &
        \num[round-precision=2,round-mode=figures,scientific-notation=true]{0.000238431524060519} &
        \num[round-precision=2,round-mode=figures,scientific-notation=true]{0.000357058643607643} \\
        Kendall $\tau$ coefficient &
        \num[round-precision=2,round-mode=figures,scientific-notation=false]{-0.0666666701436043} &
        \num[round-precision=2,round-mode=figures,scientific-notation=false]{0.333333343267441} &
        \num[round-precision=2,round-mode=figures,scientific-notation=false]{0.600000023841858} &
        \num[round-precision=2,round-mode=figures,scientific-notation=false]{0.422222226858139} &
        \num[round-precision=2,round-mode=figures,scientific-notation=false]{0.155555561184883} &
        \num[round-precision=2,round-mode=figures,scientific-notation=false]{0.422222226858139} \\
        \bottomrule
    \end{adjustbox}
\end{table}

\subsection{Steady Navier--Stokes equations}

\begin{table}[]
    \centering
    \caption{Details of neural network architectures.} \label{tb:architecture}
    \begin{tabular}{l l l l l}
    \toprule 
    Architecture & 1 & 2 & 3 & 4 \\
    \cmidrule(r){1-1} \cmidrule(r){2-2} \cmidrule(r){3-3} \cmidrule(r){4-4} \cmidrule(r){5-5}
    Hidden layers velocity network & 1 & 2 & 2 & 3 \\
    Hidden layers pressure network & 1 & 1 & 2 & 2 \\
    Number of parameters & 147 & 419 & 691 & 963 \\
    \bottomrule
    \end{tabular}
\end{table}
The steady Navier--Stokes equations are a widely used model for incompressible steady flows. In contrast to the Poisson problem, we have a system of partial differential equations
\begin{equation} \label{eq:nse}
   \begin{alignedat}{2}
    -\nu \nabla^2 \bm{u} + \bm{u} \cdot \nabla \bm{u} + \nabla p &= \bm{f} \quad & &\text{in } \Omega, \\
    \nabla \cdot \bm{u} &= 0 \quad & &\text{in } \Omega, \\
    \bm{u} &= \bm{g} \quad & &\text{on } \partial \Omega.
   \end{alignedat}
\end{equation}
Here, $\bm{u}(\bm{x}) = \left(u_1(\bm{x}) , u_2(\bm{x}) \right)$ and $p(\bm{x}) $ represent velocity and pressure, respectively. In this work, we consider the analytical solution of system~\ref{eq:nse} from~\citet{kovasznay1948laminar} on the domain ${\Omega = [-0.5, 1.0] \times [-0.5, 1.5]}$. The solution to the Kovasznay problem is given as
\begin{align}
    u_1(x_1, x_2) &= 1 - \exp \left( \lambda x_1 \right) \cos 2 \pi x_2, \\
    u_2(x_1, x_2) &= \frac{\lambda}{2\pi} \exp \left( \lambda x_1 \right) \sin 2\pi x_2, \\
    p(x_1, x_2) &= \frac{1}{2} \left( 1 - \exp \left( 2 \lambda x_1 \right)\right) + C,
\end{align}
with $\lambda = 1/(2\nu) - \sqrt{ 1/\left(4\nu^2\right) + 4\pi^2}$, $\nu = 0.025$, and $\bm{f} = \bm{0}$. Note that the pressure is just defined up to an arbitrary constant $C$. We use the analytical solution above as our boundary condition, i.e.,  $\bm{u} \rvert_{\partial \Omega} = \bm{g}$. We propose the following loss function
\begin{align}
\begin{split}
    \sqrt{L^{\text{ns}}_{MC}(\hat{\bm{u}}, \hat{p}; \theta)} &= \sqrt{L_{\text{mom}}(\hat{\bm{u}}, \hat{p}) + L_{\text{div}}(\hat{\bm{u}}) + L_{\text{bou}}(\hat{\bm{u}})}, \\
    L_{\text{mom}}\left(\hat{\bm{u}}, \hat{p} \right) &= \frac{1}{N_{\text{int}}} \sum_{i=1}^{N_\text{int}} \sum_{k=1}^2 \left(-\nu \nabla^2 \hat{\bm{u}} + \hat{\bm{u}} \cdot \nabla \hat{\bm{u}} + \nabla \hat{p}\right)_k^2 \Big\rvert_{\bm{x}_i}, \\
    L_{\text{div}}\left(\hat{\bm{u}} \right) &= \frac{1}{N_\text{int}} \sum_{i=1}^{N_\text{int}} \left(\nabla \cdot \hat{\bm{u}}\right)^2 \Big\rvert_{\bm{x}_i}, \\
    L_{\text{bou}}\left(\hat{\bm{u}} \right) &= \frac{1}{N_\text{bou}} \sum_{j=1}^{N_\text{bou}} \sum_{k=1}^2 \left( \hat{\bm{u}}-\bm{g}\right)_k^2 \Big\rvert_{\bm{s}_j}.
\end{split}
\end{align}
Note that we use two separate neural networks for $\bm{\hat{u}} = (\hat{u}_1, \hat{u}_2)$ and $\hat{p}$; alternatively we could have trained just one neural network with three outputs. Even though the latter approach might be better in capturing the underlying differential equation, we decided to use the former as it lets us approximate $\hat{\bm{u}}$ and $\hat{p}$ with neural networks of different size; this is beneficial as we are often more interested in $\hat{\bm{u}}$ than in $\hat{p}$. 

In this work, we trained four different network architectures on three datasets with 4000, 8000, and 16000 interior as well as boundary data points, respectively; the details of the architectures can be found in Table~\ref{tb:architecture} (each hidden layer has again 16 hidden units). Numerical results for these combinations can be found in Table~\ref{tb:kovasznay_1}; architectures 2-4 achieve good results for $\sum_{i=1}^2 \sqrt{FD\left( u_i-\hat{u}_i \right)}$. 

Statistics for the Kovasznay flow problem can be found in Table~\ref{tb:kovasznay_2}, where 
$FD_{\text{mom}} = FD \left(\sqrt{\sum_{k=1}^2 \left(-\nu \nabla^2 \hat{\bm{u}} + \hat{\bm{u}} \cdot \nabla \hat{\bm{u}} + \nabla \hat{p}\right)_k^2} \right)$ and $FD_{\text{div}} = FD \left( \nabla \cdot \hat{\bm{u}} \right)$.

Here we also computed the averaged error 
\begin{align}
   \widetilde{\mu} = \sum_{i=1}^2 \sqrt{FD\left(u_i - \frac{1}{10} \sum_{k=1}^{10} \hat{u}_i^{(k)} \right)},
\end{align}
over all ten networks $\hat{u}^{(1)}, \cdots, \hat{u}^{(10)}$. Note that for the combination of architecture~3 and dataset~1, $\widetilde{\mu}$ is almost two thirds smaller than $\sum_{i=1}^2 \sqrt{FD\left( u_i-\hat{u}_i \right)}$; generally the former quantity is smaller than the latter for all six combinations of architectures and datasets. The Kendall $\tau$ coefficient is $\geq 0.85$ (for the two quantities $\sum_{i=1}^2 \sqrt{FD\left( u_i-\hat{u}_i \right)}$ and $\sqrt{FD_{\text{mom}}} + \sqrt{FD_{\text{div}}}$)  for five out of the six tested combinations. 

\begin{table}[]
    \centering
    \caption{Numerical results for the Kovasznay flow problem.}
    \label{tb:kovasznay_1}
    \begin{adjustbox}{tabular=lllllll,center}
        \toprule
        Dataset & 1 & 2 & 3 & 1 & 2 & 3\\
        \cmidrule(r){1-1} \cmidrule(r){2-2} \cmidrule(r){3-3} \cmidrule(r){4-4} \cmidrule(r){5-5} \cmidrule(r){6-6} \cmidrule(r){7-7}
        Architecture & \multicolumn{3}{c}{1} & \multicolumn{3}{c}{2} \\
        \cmidrule(r){1-1} \cmidrule(r){2-4} \cmidrule(r){5-7}
        $\sum_{i=1}^2 \sqrt{FD\left( u_i-\hat{u}_i \right)}$ &
        \num[round-precision=2,round-mode=figures,scientific-notation=true]{0.014954820146163} &
        \num[round-precision=2,round-mode=figures,scientific-notation=true]{0.0138218449150613} &
        \num[round-precision=2,round-mode=figures,scientific-notation=true]{0.00807428507389834} &
        \num[round-precision=2,round-mode=figures,scientific-notation=true]{0.0000106450579786253} &
        \num[round-precision=2,round-mode=figures,scientific-notation=true]{0.00000572593250416847}  \cellcolor[gray]{.8}&
        \num[round-precision=2,round-mode=figures,scientific-notation=true]{0.00000642644157528028} \\
        $\sqrt{FD_{\text{mom}}} + \sqrt{FD_{\text{div}}}$ &
        \num[round-precision=2,round-mode=figures,scientific-notation=true]{0.0317571338750586} &
        \num[round-precision=2,round-mode=figures,scientific-notation=true]{0.0327120798774393} &
        \num[round-precision=2,round-mode=figures,scientific-notation=true]{0.019046720670454} &
        \num[round-precision=2,round-mode=figures,scientific-notation=true]{0.0000771163807786579} &
        \num[round-precision=2,round-mode=figures,scientific-notation=true]{0.0000545342557220354} &
        \num[round-precision=2,round-mode=figures,scientific-notation=true]{0.0000493450340210977} \\
        \# Iterations & $<6000$ & $<6000$ & $<8000$ & $<12000$ & $<14000$ & $<11000$ \\
        \midrule
        Architecture & \multicolumn{3}{c}{3} & \multicolumn{3}{c}{4} \\
        \cmidrule(r){1-1} \cmidrule(r){2-4} \cmidrule(r){5-7}
        $\sum_{i=1}^2 \sqrt{FD\left( u_i-\hat{u}_i \right)}$ &
        \num[round-precision=2,round-mode=figures,scientific-notation=true]{0.0000150093157090305} &
        \num[round-precision=2,round-mode=figures,scientific-notation=true]{0.0000168248287792088} &
        \num[round-precision=2,round-mode=figures,scientific-notation=true]{0.00000967062662733042} &
        \num[round-precision=2,round-mode=figures,scientific-notation=true]{0.00000636463430116495} &
        \num[round-precision=2,round-mode=figures,scientific-notation=true]{0.0000120930124234172} &
        \num[round-precision=2,round-mode=figures,scientific-notation=true]{0.0000140687157303457} \\
        $\sqrt{FD_{\text{mom}}} + \sqrt{FD_{\text{div}}}$ &
        \num[round-precision=2,round-mode=figures,scientific-notation=true]{0.0000948084731242797} &
        \num[round-precision=2,round-mode=figures,scientific-notation=true]{0.000220218643271229} &
        \num[round-precision=2,round-mode=figures,scientific-notation=true]{0.0000674140859522865} &
        \num[round-precision=2,round-mode=figures,scientific-notation=true]{0.0000472372482436197} &
        \num[round-precision=2,round-mode=figures,scientific-notation=true]{0.0000716907206900754} &
        \num[round-precision=2,round-mode=figures,scientific-notation=true]{0.000100047886084844} \\
        \# Iterations & $<12000$ & $<11000$ & $<11000$ & $<10000$ & $<8000$ & $<6000$ \\
        \bottomrule
    \end{adjustbox}
\end{table}

\begin{table}[]
    \centering
    \caption{Statistics of numerical results for the Kovasznay flow problem.}
    \label{tb:kovasznay_2}
    \begin{adjustbox}{tabular=lllllll,center}
        \toprule
        Architecture & \multicolumn{3}{c}{2} & \multicolumn{3}{c}{3} \\
        \cmidrule(r){1-1} \cmidrule(r){2-4} \cmidrule(r){5-7}
        Dataset & 1 & 2 & 3 & 1 & 2 & 3\\
        \midrule
        $\mu_{\sum_{i=1}^2 \sqrt{FD\left( u_i-\hat{u}_i \right)}}$ &
        \num[round-precision=2,round-mode=figures,scientific-notation=true]{0.0000372792889605631} &
        \num[round-precision=2,round-mode=figures,scientific-notation=true]{0.0000216575772487449} &
        \num[round-precision=2,round-mode=figures,scientific-notation=true]{0.0000374852217791861} &
        \num[round-precision=2,round-mode=figures,scientific-notation=true]{0.0000184024518398719} &
        \num[round-precision=2,round-mode=figures,scientific-notation=true]{0.0000271536306554948} &
        \num[round-precision=2,round-mode=figures,scientific-notation=true]{0.00002137243972126} \\
        $\sigma_{\sum_{i=1}^2 \sqrt{FD\left( u_i-\hat{u}_i \right)}}$ &
        \num[round-precision=2,round-mode=figures,scientific-notation=true]{0.0000329752130740883} &
        \num[round-precision=2,round-mode=figures,scientific-notation=true]{0.0000147533463091551} &
        \num[round-precision=2,round-mode=figures,scientific-notation=true]{0.0000374719631576038} &
        \num[round-precision=2,round-mode=figures,scientific-notation=true]{0.0000099908941203938} &
        \num[round-precision=2,round-mode=figures,scientific-notation=true]{0.0000157912471733615} &
        \num[round-precision=2,round-mode=figures,scientific-notation=true]{0.0000188169265076689} \\
        $\mu_{\sqrt{FD_{\text{mom}}} + \sqrt{FD_{\text{div}}}}$ &
        \num[round-precision=2,round-mode=figures,scientific-notation=true]{0.000197655985019878} &
        \num[round-precision=2,round-mode=figures,scientific-notation=true]{0.000139199945334357} &
        \num[round-precision=2,round-mode=figures,scientific-notation=true]{0.000200174453817114} &
        \num[round-precision=2,round-mode=figures,scientific-notation=true]{0.000107480744338312} &
        \num[round-precision=2,round-mode=figures,scientific-notation=true]{0.000153997841094415} &
        \num[round-precision=2,round-mode=figures,scientific-notation=true]{0.000106770130702845} \\
        $\sigma_{\sqrt{FD_{\text{mom}}} + \sqrt{FD_{\text{div}}}}$ &
        \num[round-precision=2,round-mode=figures,scientific-notation=true]{0.000124777251805716} &
        \num[round-precision=2,round-mode=figures,scientific-notation=true]{0.0000749961275921354} &
        \num[round-precision=2,round-mode=figures,scientific-notation=true]{0.000125412874204815} &
        \num[round-precision=2,round-mode=figures,scientific-notation=true]{0.0000344244784079529} &
        \num[round-precision=2,round-mode=figures,scientific-notation=true]{0.0000934563935308766} &
        \num[round-precision=2,round-mode=figures,scientific-notation=true]{0.000045919403260753} \\
        $\widetilde{\mu}$ &
        \num[round-precision=2,round-mode=figures,scientific-notation=true]{1.2884491487884412e-05} &
        \num[round-precision=2,round-mode=figures,scientific-notation=true]{7.894514686941781e-06} &
        \num[round-precision=2,round-mode=figures,scientific-notation=true]{2.0722812815155232e-05} &
        \num[round-precision=2,round-mode=figures,scientific-notation=true]{6.646698095248834e-06}  \cellcolor[gray]{.8}&
        \num[round-precision=2,round-mode=figures,scientific-notation=true]{9.248949596725163e-06} &
        \num[round-precision=2,round-mode=figures,scientific-notation=true]{8.469919622522042e-06} \\
        $\tau$ &
        \num[round-precision=2,round-mode=figures,scientific-notation=false]{0.853986442089081} &
        \num[round-precision=2,round-mode=figures,scientific-notation=false]{0.955555558204651} \cellcolor[gray]{.8}&
        \num[round-precision=2,round-mode=figures,scientific-notation=false]{0.911111116409302} &
        \num[round-precision=2,round-mode=figures,scientific-notation=false]{0.600000023841858} &
        \num[round-precision=2,round-mode=figures,scientific-notation=false]{1} \cellcolor[gray]{.8}&
        \num[round-precision=2,round-mode=figures,scientific-notation=false]{0.866666674613953} \\
        \bottomrule
    \end{adjustbox}
\end{table}

\begin{figure}[h]
    \begin{subfigure}{.5\textwidth}
      \centering
      \includegraphics[width=\linewidth]{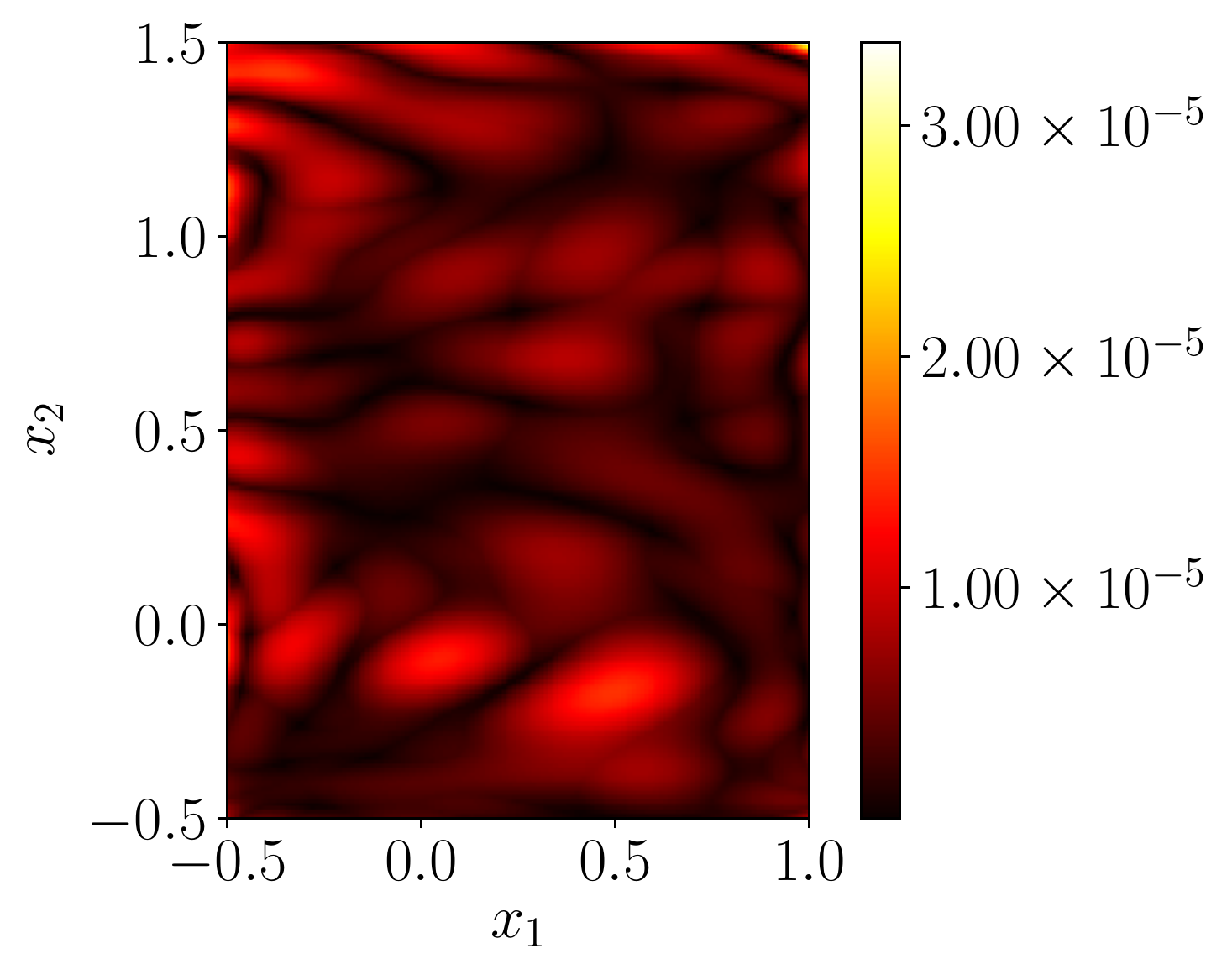}
      \caption{$\sum_{i=1}^2 \abs{u_i(\bm{x})-\hat{u}_i(\bm{x})}$}
      \label{fig:sub1}
    \end{subfigure}%
    \begin{subfigure}{.5\textwidth}
      \centering
      \includegraphics[width=\linewidth]{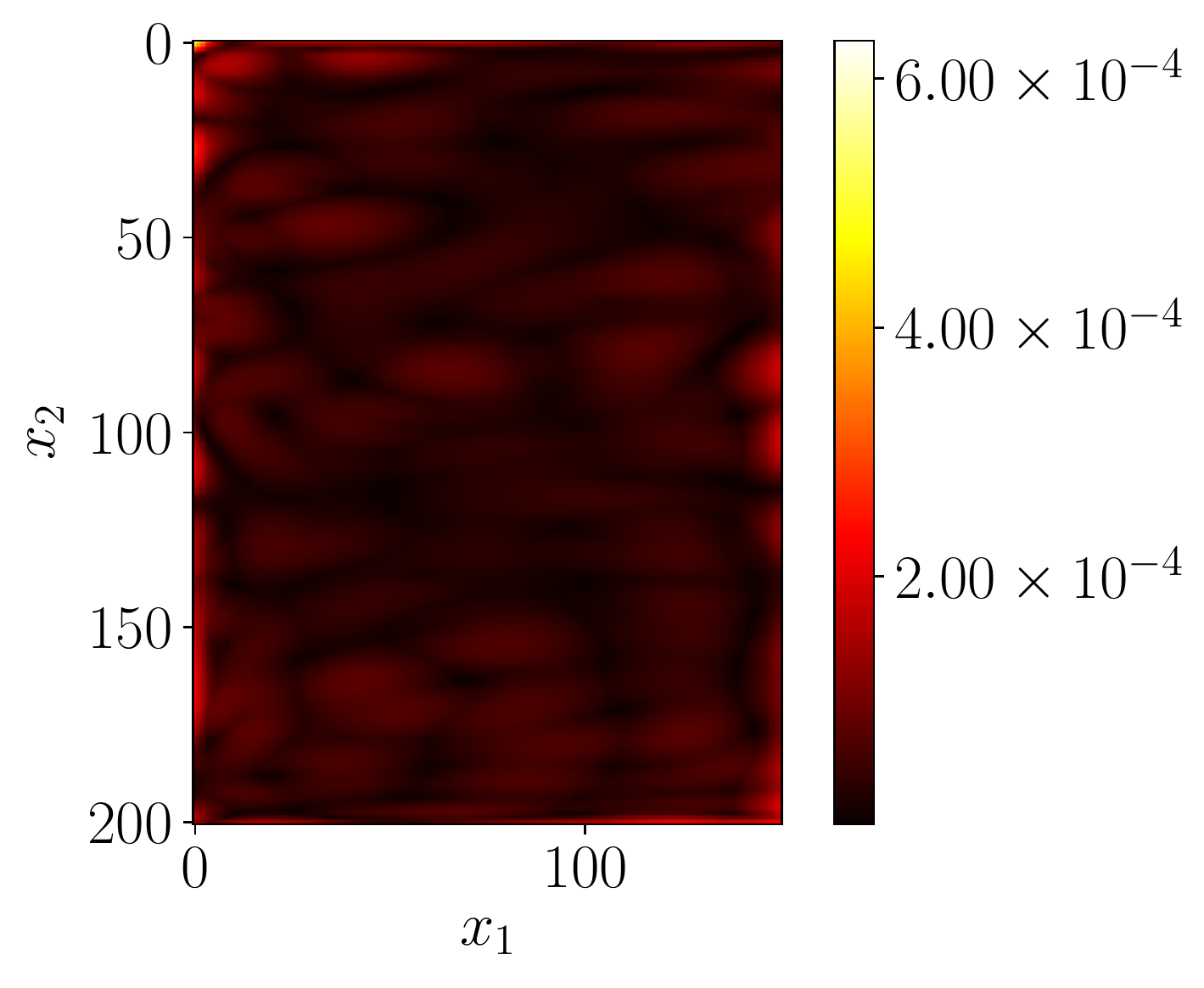}
      \caption{$\sum_{i=1}^2 \abs{\left(-\nu \nabla^2 \hat{\bm{u}} + \hat{\bm{u}} \cdot \nabla \hat{\bm{u}} + \nabla \hat{p}\right)_i} \rvert_{\bm{x}} + \abs{ \nabla \cdot \hat{\bm{u}}} \rvert_{\bm{x}}$}
      \label{fig:sub2}
    \end{subfigure}
    \caption{Numerical solution of the Kovasznay flow using architecture 2 and dataset 2.}
\end{figure}

\section{Discussion and future work}
We showed that (small) neural networks are able to learn complex solutions of (systems of) partial differential equations when optimized with the BFGS algorithm. The best results, i.e. the smallest difference of the true solution and the the approximate solution, for the manufactured Poisson problem and the Kovasznay problem are achieved with neural networks having 65 and 419 learnable parameters, respectively.

At the moment, it is unclear why the correlation of $\sqrt{FD(u-\hat{u})}$ and $\sqrt{FD(\nabla^2 \hat{u} + f)}$ for the Poisson problem is small in comparison to the correlation of $\sum_{i=1}^2 \sqrt{FD\left( u_i-\hat{u}_i \right)}$ and $\sqrt{FD_{\text{mom}}} + \sqrt{FD_{\text{div}}}$ for the Navier--Stokes problem. A strong correlation of $\sqrt{FD(u-\hat{u})}$ and $\sqrt{FD(\nabla^2 \hat{u} + f)}$ is desirable as we optimize $\sqrt{L^{\text{p}}_{MC}}$, a strongly related quantity to $\sqrt{FD(\nabla^2 \hat{u} + f)}$, in order to minimize $\sqrt{FD(u-\hat{u})}$. We suppose that a modification of the loss function could improve the correlation and leave this issue subject to future work.

Changing the loss function for the Poisson problem from $\sqrt{L^{\text{p}}_{MC}}$~\eqref{eq:genereic_loss} to  $\sqrt{\bar{L}^{\text{p}}_{MC}}$~\eqref{eq:mod_loss} could not improve results; this might be due to the set of corner points being very small compared to the number of training points leading to overfitting of the neural network at the corner points. This problem might be curable by changing the modified loss function~\eqref{eq:mod_loss} to 
\begin{align}
    \sqrt{L_{MC}^{\text{p}} + \frac{\eta}{N_{\text{corner}}} \left(  \sum_{i=1}^{N_{\text{corner}}} \left( \nabla^2 \hat{u}(\bm{x}_i) + f(\bm{x}_i)\right)^2 +  \sum_{j=1}^{N_{\text{corner}}} \left(\hat{u}(\bm{s}_j)-g(\bm{s}_j)\right)^2 \right)},
\end{align}
where $0 \leq \eta \leq 1$ has to be set appropriately.

The randomness of the used algorithm can be seen as a drawback, however, we showed that averaging over models with different random initialization can lead to significantly better results. This averaging approach seems to be well-suited for our applications as good results can already be achieved with small neural networks which can be trained in parallel. Combining this approach with the method of~\citet{berg2018unified} to prevent early stopping of the BFGS solver is subject to future work.

In this work, we sampled interior and boundary points uniformly from $\Omega$ and $\partial \Omega$, respectively. However, learning the solution of a (system of) PDEs is generally more difficult in some areas than others. We studied online sampling approaches (no results shown in this paper) were we successively sampled points from areas were the approximate solution does not satisfy the PDE well, e.g. areas were $\abs{\nabla^2 \hat{u}(\bm{x}) + f(\bm{x})}$ is large for the Poisson problem. The underlying probability density function for this approach is generally complicated and Markov chain Monte Carlo methods need to be used to sample points from it. Markov chain Monte Carlo methods are inherently difficult to parallelize, and therefore slow down the method significantly. Furthermore, note that this method is just effective if there is a high correlation of the quantity sampled from and the actual quantity that needs to be minimized.

\citet{lakshminarayanan2017simple} showed that averaging methods can be used to predict uncertainty for regression and classification problems. Using a similar approach for our algorithm, we could sample points from areas with high uncertainty. Studying this approach is subject to future work.

\citet{berg2018unified} proposed a modified version of the used algorithm in this work to learn the solution of a PDE that could possibly lead to an exact representation of the boundary data by the approximate solution. It seems to be natural to study whether or not this approach can be extended to system of PDEs, e.g., the Navier--Stokes equations.

It is left to mention that the optimization procedure seems to be the major bottleneck of our algorithm as deeper networks were not able to achieve (significantly) better results than shallow networks. This might be due to converting the constrained problem to an unconstrained optimization problem with naively choosing the penalty coefficients. It might be more sophisticated to solve the constrained optimization problem
\begin{align}
    \min_{\bm{\theta}} \mathop{\mathbb{E}}_{\bm{x} \sim \mathcal{U}(\Omega)} \left[ \left( \mathcal{N} \hat{u}(\bm{x}; \bm{\theta}) - f(\bm{x}) \right)^2 \right] \quad \text{subject to } \mathop{\mathbb{E}}_{\bm{s} \sim \mathcal{U}(\partial \Omega)} \left[ \left( \hat{u}(\bm{s}; \bm{\theta}) - g(\bm{s}) \right)^2 \right] = 0.
\end{align}
using the Lagrangian descent method (see Algorithm~\ref{ag:lagrangian}). Step~3 of Algorithm~\ref{ag:lagrangian} can be solved approximately by using the BFGS solver for the loss function 
\begin{align}
    \bar{L}_{MC}(\bm{\theta}, \lambda_i) = \frac{1}{N_{\text{int}}} \sum_{i=1}^{N_{\text{int}}} \left(\mathcal{N}\hat{u}(\bm{x}_i) - f(\bm{x}_i)\right)^2 + \frac{\lambda_i}{N_{\text{bou}}} \sum_{j=1}^{N_{\text{bou}}} \left(\hat{u}(\bm{s}_j)-g(\bm{s}_j)\right)^2,
\end{align}
or $\sqrt{\bar{L}_{MC}(\bm{\theta}, \lambda_i)}$ in the case that $\bar{L}_{MC}(\bm{\theta}, \lambda_i) \ll 1$. Note that the algorithm used in this work is equivalent to Algorithm~\ref{ag:lagrangian} with $n=1$ and $\lambda_0=1$, and therefore we expect the Lagrangian descent method to perform better as we increase $n$.

\begin{algorithm}
    \caption{Lagrangian descent method for solving PDEs with neural networks.}  \label{ag:lagrangian}
    \begin{algorithmic}[1]
        \State Initialize $\bm{\theta}_0$ and $\lambda_0$ and set learning rate $\alpha$ \;
        \For{$i \gets 1$ to $n$}
            \State $\bm{\theta}_{i+1} \gets \argmin_{\bm{\theta}} \left(\mathop{\mathbb{E}}\limits_{\bm{x} \sim \mathcal{U}(\Omega)} \left[ \left( \mathcal{N} \hat{u}(\bm{x}; \bm{\theta}) - f(\bm{x}) \right)^2 \right] \quad + \lambda_i \mathop{\mathbb{E}}\limits_{\bm{s} \sim \mathcal{U}(\partial \Omega)} \left[ \left( \hat{u}(\bm{s}; \bm{\theta}) - g(\bm{s}) \right)^2 \right]\right)$
            \State $\lambda_{i+1} \gets \lambda_i - \alpha \mathop{\mathbb{E}}\limits_{\bm{s} \sim \mathcal{U}(\partial \Omega)} \left[ \left( \hat{u}(\bm{s}; \bm{\theta}_{i+1}) - g(\bm{s}) \right)^2 \right]$
        \EndFor
    \end{algorithmic}
\end{algorithm}

A summary of benefits and drawbacks of our method compared to classical numerical methods can be found in Table~\ref{tb:fem}. We hope that these benefits as well as the ideas in this section and the results in Section~\ref{se:experiments} will stimulate future work on solving (systems of) partial differential equations using neural networks.

\begin{table}[h]
    \centering
    \caption{Benefits and drawbacks of solving partial differential equations with neural networks.}
    \label{tb:fem}
    \begin{tabularx}{\linewidth}{>{\parskip1ex}X@{\kern4\tabcolsep}>{\parskip1ex}X}
    \toprule
    \textbf{Benefits} & \textbf{Drawbacks} \\
    \cmidrule(r){1-1} \cmidrule(r){2-2}
    \textit{Expressive power:} Already small neural networks are able to approximate the solution of partial differential equations well. \par 
    \textit{Ease of implementation:} Our algorithm is straightforward and can be easily implemented using backpropagation. \par
    \textit{Arbitrary domains (in higher dimensions):} The algorithm is based on drawing random points from a domain, which can be readily extended to arbitrary domains; no triangulation of the domain is needed. \par 
    \textit{Freedom of approximation spaces:} For some classical methods and systems of PDEs, we have certain restrictions on the function spaces of the solutions, e.g., the inf--sup condition \cite{brezzi2012mixed} needs to be satisfied for the Navier--Stokes equations when using the finite element method. \par
    \textit{Sensor data:} We can easily incorporate (noisy) information of sensors by adding a  term to the loss function.
    &
    \textit{Convergence:} Most probably due to optimization issues, we could not empirically show that errors decrease (with a certain convergence rate) with increasing neural network size; showing theoretical convergence  seems to be even more difficult. \par
    \textit{Run time:} At least when the BFGS optimizer is used, our simulations seem to be considerably slower than classical numerical methods. Using the proposed algorithm in combination with a first order optimization method, e.g. Adam~\cite{kingma2014adam}, is subject to future work. \par
    \textit{Scalability:} At the moment, it is unclear how well the algorithm will scale to more difficult problems, e.g., three-dimensional turbulent flows. \par 
    \textit{Randomness:} Different random initializations lead to different results of the algorithm whereas (most) classical numerical methods are deterministic. Further investigation is necessary to better exploit this non-determinism and to eventually turn it into an advantage.
    \\ \bottomrule
    \end{tabularx}
\end{table}

\newpage
\bibliographystyle{unsrtnat}
\bibliography{main}

\end{document}